\documentclass[a4paper,conference]{IEEEtran}
\IEEEoverridecommandlockouts
\usepackage{cite}
\usepackage{amsmath,amssymb,amsfonts}
\usepackage{algorithmic}
\usepackage[pdftex]{graphicx}
\usepackage{textcomp}
\usepackage{xcolor}
\usepackage{hyperref}
\hypersetup{
    colorlinks=true,
    linkcolor=blue,
    filecolor=magenta,      
    urlcolor=cyan,
}

\usepackage{amsmath,amsfonts,bm}

\def\eqref#1{equation~\ref{#1}}

\def\1{\bm{1}}

\DeclareMathAlphabet{\mathsfit}{\encodingdefault}{\sfdefault}{m}{sl}
\SetMathAlphabet{\mathsfit}{bold}{\encodingdefault}{\sfdefault}{bx}{n}

\usepackage{booktabs}       
\usepackage{tabularx}
\usepackage{multirow}
\usepackage{arydshln}

\def\BibTeX{{\rm B\kern-.05em{\sc i\kern-.025em b}\kern-.08em
    T\kern-.1667em\lower.7ex\hbox{E}\kern-.125emX}}
\begin{document}

\title{Improving Visual Relation Detection \\ using Depth Maps}

\makeatletter
\newcommand{\linebreakand}{%
  \end{@IEEEauthorhalign}
  \hfill\mbox{}\par
  \mbox{}\hfill\begin{@IEEEauthorhalign}
}
\makeatother

\author{\IEEEauthorblockN{Sahand Sharifzadeh}
\IEEEauthorblockA{\textit{Ludwig Maximilian University of Munich}\\
sharifzadeh@dbs.ifi.lmu.de}
\and
\IEEEauthorblockN{Sina Moayed Baharlou}
\IEEEauthorblockA{\textit{Sapienza University of Rome}\\
baharlou@dis.uniroma1.it}
\and
\IEEEauthorblockN{Max Berrendorf}
\IEEEauthorblockA{\textit{Ludwig Maximilian University of Munich}\\
berrendorf@dbs.ifi.lmu.de}
\linebreakand 
\IEEEauthorblockN{Rajat Koner}
\IEEEauthorblockA{\textit{Ludwig Maximilian University of Munich}\\
koner@dbs.ifi.lmu.de}
\and
\IEEEauthorblockN{Volker Tresp}
\IEEEauthorblockA{\textit{Ludwig Maximilian University of Munich} \\ \textit{\& Siemens AG}\\
volker.tresp@siemens.com}
}

\maketitle

\begin{abstract}
Visual relation detection methods rely on object information extracted from RGB images such as 2D bounding boxes, feature maps, and predicted class probabilities. We argue that depth maps can additionally provide valuable information on object relations, e.g. helping to detect not only spatial relations, such as \texttt{standing behind}, but also non-spatial relations, such as \texttt{holding}. In this work, we study the effect of using different object features with a focus on depth maps. To enable this study, we release a new synthetic dataset of depth maps, \textit{VG-Depth}, as an extension to Visual Genome (VG). We also note that given the highly imbalanced distribution of relations in VG, typical evaluation metrics for visual relation detection cannot reveal improvements of under-represented relations. To address this problem, we propose using an additional metric, calling it \textit{Macro Recall@K}, and demonstrate its remarkable performance on VG. Finally, our experiments confirm that by effective utilization of depth maps within a simple, yet competitive framework, the performance of visual relation detection can be improved by a margin of up to $8\%$.
\end{abstract}


\section{Introduction}
\label{introduction}
Scene Graph Generation, i.e. detecting objects and their relations in images in form of \texttt{(subject, predicate, object)}, is a fundamental task in scene understanding and can play an important role in  recommender systems, visual question answering, decision making, etc. For example, detecting whether a man is \texttt{on} a bike or \texttt{next to} a bike is a crucial challenge in autonomous driving. Most works in this area rely on image-based object information such as class labels, bounding boxes and RGB features. We argue that depth maps can additionally provide valuable information about an object's relations as they provide the objects' distance from the camera. This information can help to distinguish between many relations such as \texttt{behind}, \texttt{in front of} and even improve detection in situations where the objects are nearby such as \texttt{covered in}. Figure \ref{fig:example} shows a successfully detected example of the relation \texttt{(fence, behind, dog)} after employing its depth map, and using our model. The goal of this work is to study the effect of using different object features on visual relation detection, with a focus on depth maps.

\begin{figure}
        \begin{center}
        \includegraphics[width=1.\linewidth]{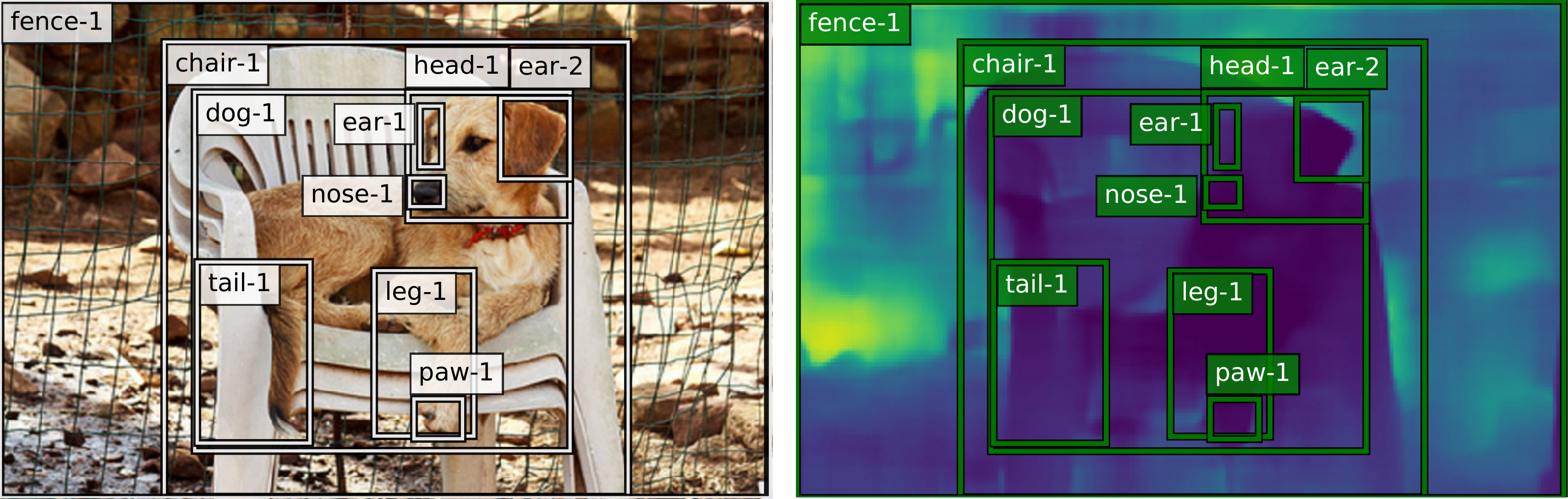}
        \end{center}
         \caption{An image from the VG dataset (left), and the corresponding synthetically generated depth map from VG-Depth dataset (right), annotated by the scene graph. Bright colors in the depth map indicate a larger distance to the camera. Utilizing depth maps allows us to successfully predict the relation \texttt{(fence-1, behind, dog-1)}.}
         \label{fig:example}
\end{figure}

Unfortunately, most available image datasets, specifically the ones with relational annotations such as Visual Relation Detection (VRD) \cite{lu2016visual} and Visual Genome (VG) \cite{krishna2017visual}, do not provide depth maps, because the acquisition of depth maps is a cumbersome task requiring specialized hardware. We tackle this issue by synthetically generating the corresponding \textit{pseudo} depth maps from 2D images of Visual Genome. This is possible thanks to the large corpora of available RGB-D pairs, i.e. NYU-Depth-v2 \cite{Silberman:ECCV12} dataset. Using RGB-D pairs in NYU-Depth-v2 and a fully convolutional neural network, allow us to learn the mapping function of RGB images to their corresponding depth maps. We can then apply this network to the images from VG, generating their corresponding depth maps. We release the depth maps that are generated from VG, as an extention to it, calling it \textbf{\emph{VG-Depth}}\footnote{The dataset and our framework are publicly available at \url{https://github.com/Sina-Baharlou/Depth-VRD}.}. The object information extracted from depth maps and RGB images, i.e. class labels, location vectors, RGB and depth features, are the basis for relation detection in our simple yet effected framework.

Additionally, we note that the typically employed Recall@K metric (Micro Recall@K), cannot properly reveal the improvements of under-represented relations in highly imbalanced datasets such as VG. This might be an issue in applications such as autonomous driving where it is important to ensure that the model is capable of predicting also important but less represented predicates such as \textit{walking on} (648 in VG test set) and not just \textit{wearing} (20,148 in VG test set). We address this issue by proposing to employ \textbf{\textit{Macro Recall@K}}, where we compute the mean over Micro Recall@K per predicate, thereby eliminating the effect that over-represented classes have in Micro Recall@K setting.

In summary, our contributions are as follows: 
\begin{enumerate}
\item We perform an extensive study on the effect of using different sources of object information in visual relation detection. We show in our empirical evaluations using the VG dataset, that our model can outperform competing methods by a margin of up to $8\%$ points, even those using external language sources or contextualization.
\item We release a new synthetic dataset \textit{VG-Depth}, to compensate for the lack of depth maps in Visual Genome.
\item We propose \textit{Macro Recall@K} as a competitive metric for evaluating the visual relation detection performance in highly imbalanced datasets such as Visual Genome.
\end{enumerate}

\section{Related Works}
\label{related}
\paragraph{Knowledge Graph (KG) Modeling} In Knowledge Graph modeling, the aim is typically to find embeddings or latent representations for entities and predicates, which then can serve to predict the probability of unseen triples. These methods mostly differ in how they model relations. In RESCAL~\cite{nickel2011three} each relation is defined as a transformation in the embedding space of entities, producing a triple probability. TransE~\cite{NIPS2013_5071} employs a similar idea but limits each relation to a translation. In comparison to RESCAL, it has fewer parameters; as a disadvantage, it cannot model symmetric relations. DistMult~\cite{yang2014embedding} considers each relation as a vector,  similar to TransE,  but minimizes the trilinear dot product of subject, predicate and object vector. DistMult can be understood as a form of RESCAL, where the transformation matrix is diagonal. ComplEx~\cite{trouillon2016complex} extends DistMult to complex-valued vectors of embeddings. A multilayer perceptron (MLP) architecture~\cite{dong2014knowledge} extends these methods to non-linear transformations and has shown to be competitive to the other discussed approaches on most benchmarks~\cite{nickel2016review,socher2013reasoning}. For an extensive review and study on different KG models refer to~\cite{nickel2016review, ali2020bringing, ali2020pykeen}.

\paragraph{Scene Graph (SG) Generation} SG Generation started with the release of Visual Relation Detection (VRD)~\cite{lu2016visual} and the VG~\cite{krishna2017visual}.
In VRD, Word2Vec representations of the subject, object, and the predicate were used to train a model jointly with the corresponding image region that describes the predicate. In particular, they consider the joint bounding box of subject and object as the image representation for the predicate. Follow-up work achieved improved performance by incorporating a knowledge graph, constructed from the image annotations~\cite{baier2017improving}.
Later, VTransE employed TransE~\cite{bordes2013translating} to model visual relations. More recently, Yu et al. ~\cite{yu2017visual} proposed a teacher-student model to distill external language knowledge to improve visual relation detection. Iterative Message Passing~\cite{xu2017scene}, Neural Motifs~\cite{zellers2018neural} (NM) and Graph R-CNN~\cite{yang2018graph} incorporate context within each prediction using RNNs and graph convolutions respectively. For an extensive discussion on the connection between scene graphs and knowledge graphs refer to~\cite{tresp2019model,tresp2020tensor}.

\paragraph{Depth Maps} While several works have leveraged depth maps to improve \textit{object} detection~\cite{bo2013unsupervised,eitel2015multimodal,gupta2014learning}, the idea of using depth maps in the \textit{relation} detection task has only been explored recently: Yang et al.~\cite{yang2018visual} employ a basic framework for visual relation detection, with handcrafted depth map features, i.e. the mean and mode over pixel values of each depth map. They have a limited experimental setting, where they consider only human-centered relations. In this work, we explore the usability of depth maps in a larger domain and using a convolutional neural network for feature extraction. Furthermore, we provide a more extensive study, release a relevant dataset, and propose a more suitable metric.

\section{Framework}
\label{sec_models}
In this section, we introduce the framework that we employed for this study. Let $\mathcal{E} = \{e_1, e_2, ..., e_n\}$ be the set of all entities, including subjects ($s$) and objects ($o$), and $\mathcal{P} = \{p_1, p_2, ..., p_m\}$ the set of all predicates. Each entity $e_i$ can appear in images within a bounding box $\mathbf{bb_i}=[x_i,y_i,w_i,h_i]$, from an image $\mathbf{I}$, where $[x_i,y_i]$ are the coordinates of the bounding box and $[w_i,h_i]$ are its width and height. In this work we apply Faster R-CNN~\cite{ren2015faster}, on each image $\mathbf{I}$ to extract a feature map $\mathbf{fmap_I}$, together with object proposals as a set of bounding boxes $\mathbf{bb}$ and class probability distributions $\mathbf{c}$. For each RGB image, we generate a depth map $\mathbf{D}$ where the same bounding box areas encompass the entities' distance from the camera. In the next section, we first describe the synthetic generation of $\mathbf{D}$s and then the feature extraction from generated depth maps. In the end, we describe the relation detection module, where the pairwise features are fused and then employed for relation detection.

\begin{figure*}
    \begin{center}
      \includegraphics[width=.65\textwidth]{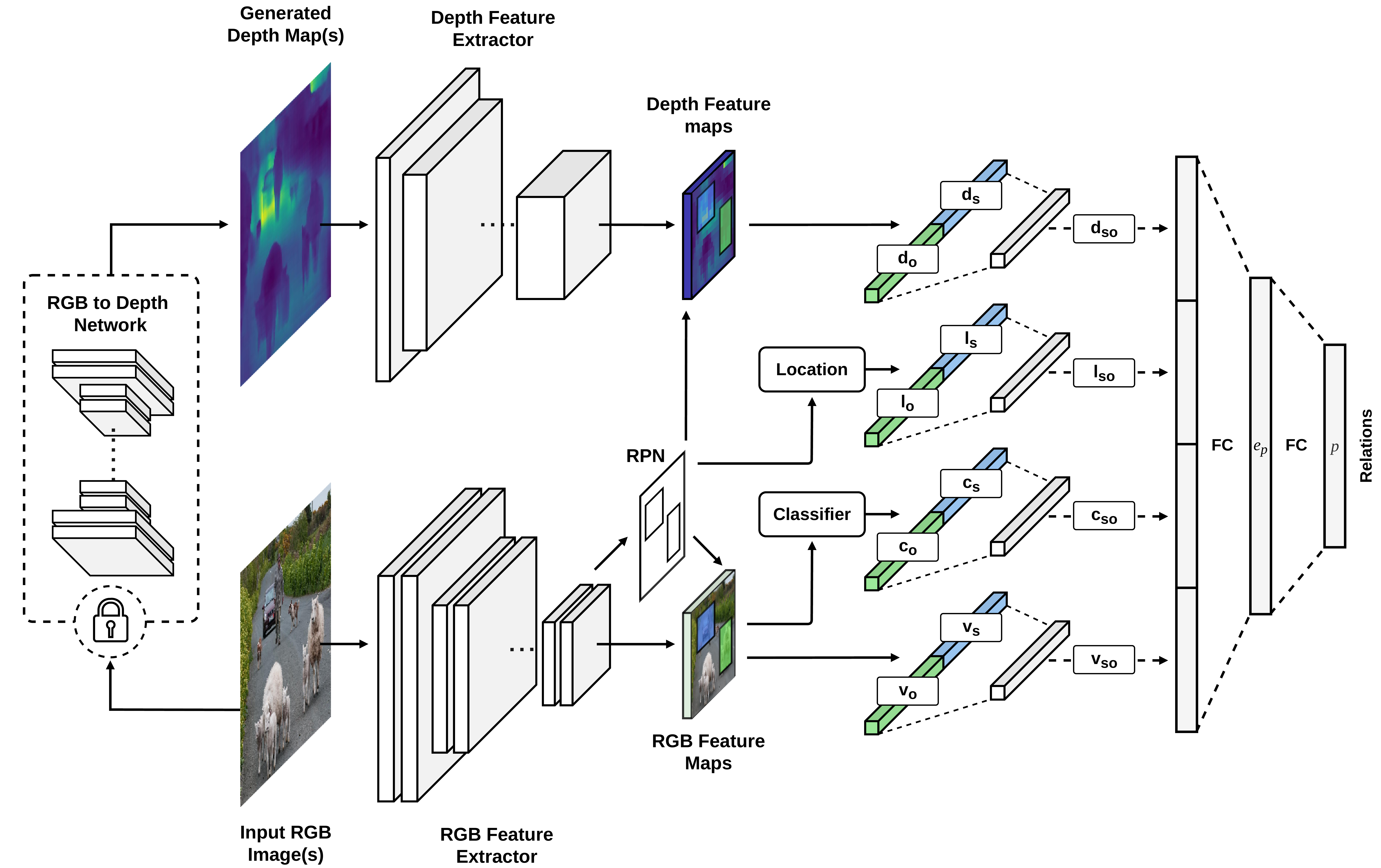}
      \end{center}
  \caption{We study the effect of object information, i.e. class labels, location vectors, RGB and depth features in visual relation detection by employing the simple yet effective framework presented in this figure. We generate depth maps synthetically using an RGB-to-Depth model, eliminating the need for specialized hardware. On the left side, we see the RGB image and its generated depth map, fed into CNNs to extract feature maps from both modalities. We create pairwise feature vectors $\mathbf{d_{so}}$ (pooled from depth feature maps), $\mathbf{l_{so}}$ (from bounding boxes), $\mathbf{c_{so}}$ (from class labels) and $\mathbf{v_{so}}$ (pooled from RGB features) and feed them into a relation detection layer to infer the predicate.
 }\label{models_img}
\end{figure*} 

\subsection{Depth Maps for Relation Detection}
\subsubsection{Generation}\label{generation}
We incorporate an RGB-to-Depth model within our visual relation detection framework. As shown in Figure \ref{models_img}, this is a fully convolutional neural network (CNN) that takes an RGB image as input and generates its predicted depth map. This model can be pre-trained on any datasets containing pairs of RGB and depth maps regardless of having the class annotations for objects or predicates. This enables us to work with the already available visual relation detection datasets without requiring to collect additional data, and also mitigates the need for specialized hardware in real-world applications. The architectural details are explained in Section~\ref{evaluation} and the generated depth maps from VG are separately released as a dataset called \textit{VG-Depth}.
\subsubsection{Feature Extraction}\label{feature}
Depth maps have been employed in tasks such as \emph{object detection} and \emph{segmentation}~\cite{eitel2015multimodal,hazirbas2016fusenet}. In these works, it is common to simply render a depth map as an RGB image, and
extract depth features using a CNN, that has been pre-trained with RGB images (for object detection). They argue that the edges in depth maps might yield better object contours than the edges in cluttered RGB images and that one may combine edges from both RGB and depth to obtain more information~\cite{hazirbas2016fusenet}. Therefore, they aim to get similar, complementary features from both modalities.
However, the practice of employing a model pre-trained on a particular source modality, e.g. RGB, and applying it on a different target modality, e.g. depth map, is sub-optimal in many applications (one should also keep in mind that even fine-tuning some layers of a network does not change the very early convolutional filters). Hence, unlike other works, we train a feature extractor CNN directly on depth maps and specifically for the task of relation detection. Given a depth map $\mathbf{D}$, this network generates a feature map $\mathbf{fmap_{D}}$. The architectural details of this network is presented in Section \ref{evaluation}.

\subsection{Relation Model}
In the previous section, we described methods for the extraction of $\mathbf{fmap_{I}}$, $\mathbf{fmap_{D}}$, $\mathbf{c}$ and $\mathbf{bb}$. Here, we outline the model that infers relations using pairwise combinations of these features. For each pair of detected objects within an image, we create a scale-invariant location feature $\mathbf{l_{s}} = [t_x, t_y, t_w, t_h]$ with:
$
t_x = (x_s - x_o)/w_o, t_y = (y_s - y_o)/h_o, t_w = \log(w_s/w_o), t_h = \log(h_s/h_o)
$ and similarly $\mathbf{l_{o}}$. We then pool the corresponding features $\mathbf{v_s}$ and $\mathbf{v_o}$ from $\mathbf{fmap_I}$ and create a visual feature vector $[\mathbf{v_s}; \mathbf{v_o}]$. Similarly, we create a depth feature vector $[\mathbf{d_s}; \mathbf{d_o}]$, by pooling features from $\mathbf{fmap_{D}}$, within $\mathbf{bb}_s$ and $\mathbf{bb}_o$. Additionally, we create $[\mathbf{c_s}; \mathbf{c_o}]$ and $[\mathbf{l_s}; \mathbf{l_o}]$. Each of these vectors are fed into separate fully connected layers, followed by ReLUs, yielding $\mathbf{v_{so}}$, $\mathbf{l_{so}}$, $\mathbf{c_{so}}$ and $\mathbf{d_{so}}$ before being fed to the relation head which projects them to the relation space such that:
\begin{equation}\label{eq_2}
\mathbf{e_p}=f(\mathbf{W}[\mathbf{v_{so}};\mathbf{l_{so}};\mathbf{c_{so}};\mathbf{d_{so}}])
\end{equation}

Here, $\mathbf{W}$ describes a linear transformation and $f(.)$ is a non-linear function. We realize them as a fully connected layer in a neural network with ReLU activations and dropout. $\mathbf{e_p}$ is an embedding vector of pairwise features. This simple relation prediction model is inspired by the work of~\cite{dong2014knowledge} to predict links in knowledge graphs. Therefore, we call it \textbf{ERMLP-E}, short for ERMLP-Extended. The input of their proposed model is a triple and the output is a single Bernoulli variable, whereas in our work the inputs are \textit{subject} and \textit{object} and we have a Bernoulli variable for each predicate class in the output. This gives us fewer parameters compared to that model, and simplifies training by imposing an implicit negative sampling through the cross-entropy loss.

\begin{table*}[t]
    \centering
    \caption{Predicate prediction recall values on VG test set. When the depth maps are utilized together with all other features (\emph{Ours-$\mathbf{l, c, v, d}$}), we gain a large improvement compared to the state-of-the-art. One can also see that even replacing depth maps with visual features (\emph{Ours-$\mathbf{l,c,d}$} compared to \emph{Ours-$\mathbf{l, c, v}$}) can yield better results. Additionally, comparing \emph{Ours-$\mathbf{l, c, v}$} to \emph{VTransE} and \emph{Neural Motifs} reveals the advantage of our simple model regardless of depth maps.}
    \label{tab:3}
    \begin{center}
    \scalebox{1.1}{
        \begin{tabular}{cl|ccc|ccc}
            \toprule
            & Strategy         & \multicolumn{3}{c|}{\textbf{Macro} } & \multicolumn{3}{c}{\textbf{Micro}}\\
            & Task         & \multicolumn{3}{c|}{Predicate Pred.}& \multicolumn{3}{c}{Predicate Pred.}\\
            & Metric      & R@100   & R@50          & R@20    & R@100 & R@50 & R@20 \\\midrule
            \multirow{7}{*}{\rotatebox{90}{models}}
            & VTransE~\cite{zhang2017visual}        &-        & -           & -   &62.87 &62.63 &-  \\
            & Yu's-S~\cite{yu2017visual}            &-         &-       &-    & 49.88 &- &-     \\
          & Yu's-S+T~\cite{yu2017visual}            &-          &-       &- &55.89 &- &-         \\
          & IMP~\cite{xu2017scene}       &-          &-       &-   &53.00 &44.80 &-      \\
          & Graph R-CNN~\cite{yang2018graph}         &-          &-       &- &59.10 &54.20 &-          \\
          & NM~\cite{zellers2018neural}         &14.39          &13.20       &10.25 &67.10 &65.20 &58.50         \\
          \hdashline
            \multirow{9}{*}
            {\rotatebox{90}{ablations}}& 
            Ours - $d$     &\phantom{0}9.51            &\phantom{0}8.46        &\phantom{0}6.35 &54.72 &51.90 &43.86\\
            & Ours - $c$        &15.65             &13.09      &\phantom{0}8.56   &64.82 &60.54 &49.89            \\
            & Ours - $v$        &13.88             &12.24      &\phantom{0}8.99   &61.72 &58.50 &50.41            \\
            & Ours - $l$        &\phantom{0}5.19             &\phantom{0}4.66      &\phantom{0}3.57   &49.07 &46.13 &37.48            \\
            & Ours - $v, d$    &15.47           &14.04      &10.83     &62.88 &60.52 &53.07          \\
            & Ours - $l, v, d$  &15.76           &14.40      &11.07     &63.06 &60.83 &53.55    \\
            & Ours - $l, c, d$    &21.67           &19.56      &15.12     &67.97 &66.09 &59.13          \\
            & Ours - $l, c, v$    &19.16           &17.72      &13.93     &67.94 &66.06 &59.14          \\
            & Ours - $l, c, v, d$     &\textbf{22.72}       &\textbf{20.74}    &\textbf{16.40}  &\textbf{68.00}       &\textbf{66.18}    &\textbf{59.44}   \\
            \bottomrule
        \end{tabular}}\label{tab_pred}
        \end{center}
\end{table*}

As shown in earlier works, using more sophisticated models for context propagation between objects with RNNs or graph convolutions, can further improve the prediction accuracy. However, the aim here is to study the effect of including depth maps as additional object features in visual relation detection and as will be shown later, even with this simple model, utilizing depth maps can be more effective than e.g. propagating context. Clearly, those other models can also further enrich their understanding of object relations by employing depth maps.

To learn the parameters, we consider each relation \texttt{(subject, predicate, object)} with an associated Bernoulli variable that takes $1$ if the triple is observed and $0$ otherwise, following a locally closed world assumption~\cite{nickel2016review}.
Given the set of observed triples $\mathcal{T}$, the loss function is the categorical cross entropy between the one-hot targets and the distribution obtained by softmax over the network's output defined as:
\begin{equation}
\mathcal{L}=\sum\limits_{(s,p,o) \in \mathcal{T}} -\log\, \frac{\exp{(\mathbf{w'}^{\text{T}}_{p}\mathbf{e_p}})}
{\sum_{p^\prime \in \mathcal{P}} \exp{( \mathbf{w'}^{\text{T}}_{p^\prime}\mathbf{e_p})}}
\end{equation}
where $\mathbf{w'}_{p}$ is the weight vector corresponding to $p$ in the last layer (linear classification layer).

\section{Evaluation}\label{evaluation}
In our study, we are interested to answer the following questions: 
\begin{enumerate}
\item If we are given \textit{only} depth maps of some objects in a scene (and not even object labels), how accurately can we infer the distribution of possible pairwise relations? How do other sources of object information compare to it?
\item Current visual relation detection frameworks commonly rely on extensive object information such as class labels, bounding boxes, RGB features, contextual information, etc. Do depth representations bring any additional information or would they only contribute redundant scene knowledge?
\end{enumerate}
Additionally, we study whether Recall@K can sufficiently reflect the improvements of under-represented relations within a highly imbalanced dataset such as VG.

In what follows, we introduce the dataset, metrics, architectural details and experiments to answer these questions.
\subsection{Dataset}
We test our approach on the \emph{Visual Genome}~\cite{krishna2017visual} dataset. We use the more commonly used subset of VG dataset proposed by~\cite{xu2017scene} which contains 150 object classes and 50 relations.

\subsection{Metrics} 
\paragraph{Micro Recall@K} This metric is defined as the mean prediction accuracy in each image given the top $K$ predictions and is typically called \textit{Recall@K}. We assigned the \textit{Micro} prefix to its name to distinguish this metric with \textit{Macro Recall@K}. Recall@K is a popular choice in most of the visual relation detection studies. The main reason is the incompleteness of visual relation detection datasets, i.e. some relations might not be annotated in the test set, while due to the model's generalization, they might get higher prediction values than the annotated ones. This sensitivity is handled by the $K$ parameter in Recall@K.

\paragraph{Macro Recall@K} We define this metric as:
\begin{equation}
    \textsc{Macro Recall@K} = \sum_{(s,p,o) \in \mathcal{T}_p} \frac{\textsc{Micro R@K}(p)}{\vert\mathcal{T}_p\vert}
\end{equation}
where $\mathcal{T}_p \subset \mathcal{T}$ is set of all relations with predicate $p$, and $\textsc{Micro R@K}(p)$ is computed on $\mathcal{T}_p$. The motivation behind this metric is the highly imbalanced distribution of classes in some datasets such as VG. In these datasets Micro Recall@K score gets dominated by frequently labeled relations and might not reflect the improvements in some important but under-represented classes. However, in Macro R@K, the prediction accuracy of under-represented classes can have a stronger effect on the output.  This metric is inspired from the Macro F1 measure~\cite{schutze2008introduction}.

\begin{figure*}
  \begin{center}
    \includegraphics[width=1.0\textwidth]{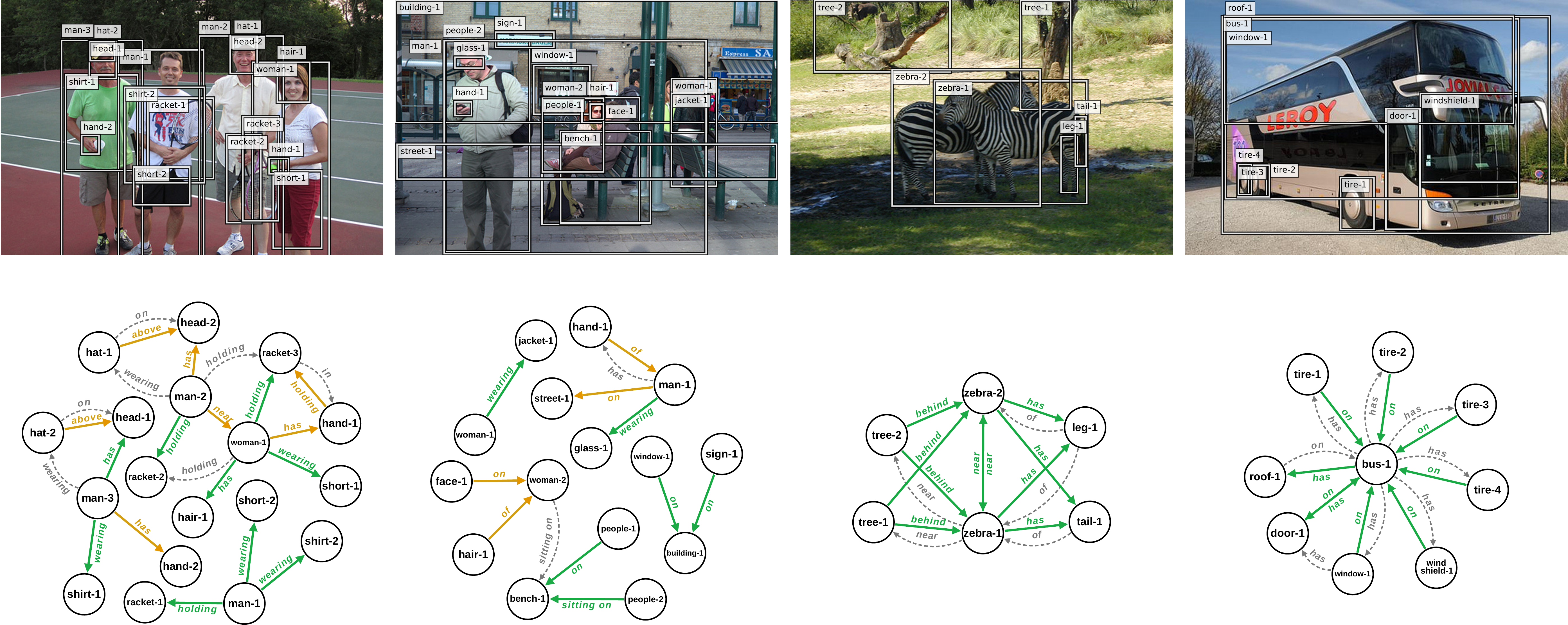}
    \end{center}
    \caption{Some of the qualitative results from our model's predictions. Green arrows indicate the successfully detected predicates (true positives), orange arrows indicate the false negatives and gray arrows indicate predicted links which are not annotated in the ground truth.}\label{fig_qualitative}
\end{figure*}

\subsection{Architectures}
\paragraph{RGB-to-Depth Network}
We employ the RGB-to-Depth architecture that has been introduced in~\cite{laina2016deeper}. The model is a fully convolutional neural network built on ResNet-50~\cite{DBLP:journals/corr/HeZRS15}, and trained in an end-to-end fashion on data from NYU Depth Dataset v2~\cite{Silberman:ECCV12}. In our experiments, we also trained the model from the outdoor images of Make3D dataset~\cite{saxena2007learning}. However, the model that was trained on this dataset, did not show promising results for relation detection. This observation is not surprising because unlike Visual Genome, Make3D images contain mostly outdoor scenes with very few objects.

\paragraph{RGB Feature Extraction} To extract embeddings and class probabilities of RGB images, we use the VGG-16 architecture~\cite{simonyan2014very} pre-trained on ImageNet~\cite{ILSVRC15} and fine-tuned on VG by Zellers et al.~\cite{zellers2018neural}.

\paragraph{Depth Map Feature Extraction}
For depth map extraction we use ResNet-18 proposed in~\cite{DBLP:journals/corr/HeZRS15}. We trained this model from scratch following the earlier discussions in Subsection~\ref{feature}. This network was trained separate from other inputs and on a pure depth-based, relation detection task using Adam~\cite{kingma2014adam}, with a learning rate of $10^{-4}$ and batch size of 32 for 30 epochs.

\paragraph{Relation Detection Network} In relation detection head, each extracted feature pair goes to a separate, fully connected hidden layer of 64 neurons ($\sim$12K learnable weights) for class probabilities, 512 for RGB feature maps ($\sim$4M learnable weights), 4096 for depth feature maps ($\sim$4M learnable weights) and 20 for location features (160 learnable weights). Each of them with a dropout rate of 0.1, 0.8, 0.6 and 0.1. The concatenated outputs are then connected to a fully connected hidden layer of 4096 neurons with 0.1 dropout and then to the classification layer. We trained this network by Adam~\cite{kingma2014adam}, with a learning rate of $10^{-5}$. We used a batch size of 16 and 30 epochs of training. All of the layers were initialized with Xavier weights~\cite{glorot2010understanding}.

\subsection{Comparing Methods}
We compare our results with \emph{VTransE}~\cite{zhang2017visual} that takes visual embeddings and projects them to relation space using TransE. We also compare to the student network of~\cite{yu2017visual} (\emph{Yu's-S}), and their full model (\emph{Yu's-S+T}) that employs external language data from Wikipedia. From the context propagating methods, we report Neural Motifs~\cite{zellers2018neural}, Graph R-CNN~\cite{yang2018graph} and IMP~\cite{xu2017scene}. In an ablation study, we report our relation prediction results under several settings in which different combinations of object information are employed for prediction.

\subsection{Experiments}\label{subsec_exp}
As our main goal is to investigate the role of depth maps and other features in relation detection, we report \textit{predicate prediction} results. In this setting, the relation detection performance is analyzed by isolating it from the object detector's error. Therefore, the goal is to evaluate the relation detection accuracy given the objects in an image. We carried on our experiments by training each model 8 times with different random seeds. The maximum variance of the results was no more than 0.01. The results are shown in Table~\ref{tab_pred}. In what follows, we provide a discussion over the quantitative and qualitative results.

\begin{figure*}
    \begin{center}
      \includegraphics[width=0.8\linewidth]{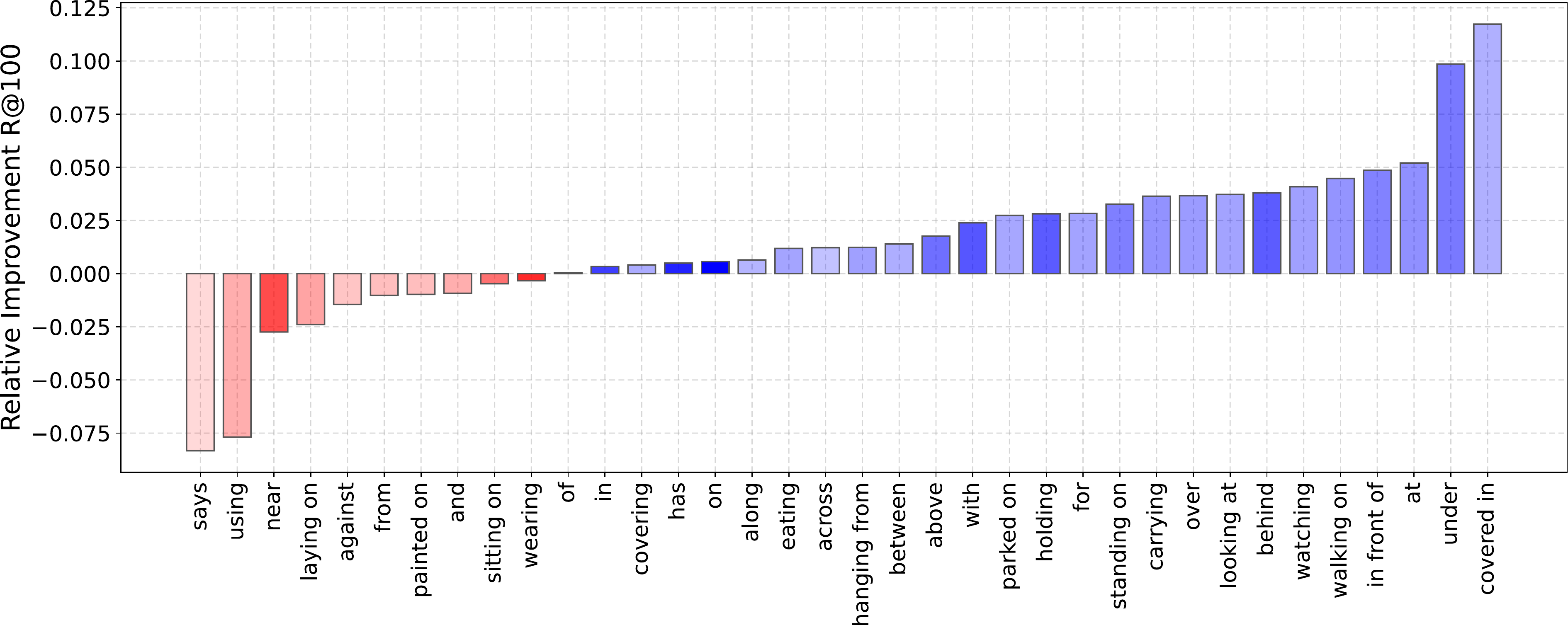}
      \end{center}
\caption{This plot shows the prediction changes per predicate, going from \emph{Ours-$v$} to \emph{Ours-$v, d$}. The classes with zero changes are omitted from the plot. The darker shades indicate larger number of that class within the test set whereas the lighter shades are under-represented classes. An improvement in predicates with more frequency has a larger effect on the Micro R@K whereas this effect is eliminated within Macro R@K. We can see that indeed the improvements by using depth maps are mostly happening within the less-represented classes.}\label{fig_bars}
\end{figure*}

The upper part of the table demonstrates the results directly reported from other works while the lower part presents the results from the ablation study on our model. For NM, we have computed the Macro R@K results using their publicly available code. We can see that our full model with depth maps, achieves the highest accuracy in comparison to the others in all settings. It is also interesting to note that when using \textit{only} depth maps we can already achieve a significant accuracy in predicate prediction, emphasizing the value of relational information that are stored within the depth maps alone. By comparing \emph{Ours-$v$} to \emph{Ours-$v, d$}, we can observe the improvements that depth maps bring. Also comparing \emph{Ours-$l, c, d$} to \emph{Ours-$l, c, v$} is specially informative from two aspects: (1) It shows that while some results are almost equal in Micro settings, one can observe a significant difference in the Macro setting, demonstrating the effectiveness of this metric in presenting the improvements of under-represented classes. (2) We observe that $v$ alone has a higher R@K than $d$ alone. However, when we add them separately to $c, l$ we can see that $d$ has more to offer. In other words, $v$ brings more redundant information to $c, l$ compared to $d$. 
To get a better intuition of the improvements that we gain after including depth maps (\emph{Ours-$v,d$} compared to \emph{Ours-$v$}), we plotted the changes in prediction accuracy for each predicate in Figure \ref{fig_bars}. We used darker shades for over-represented classes and lighter shades for under-represented ones. This helps to also gain a better intuition of improvement versus frequency of data. For example we can see that in general the accuracy of relations including the predicates such as \texttt{under}, \texttt{in front of} and \texttt{behind} has been improved. These predicates appear much less often in the dataset than \texttt{on} or \texttt{has}, having less effect in the computed \textit{Micro} accuracy. Figure \ref{goodones} presents some samples of synthetically generated depth maps in VG-Depth dataset including both high quality and faulty ones. Additionally, we present some of the predicted relations by our model in Figure \ref{fig_qualitative}.
\begin{figure*}
    \begin{center}
      \includegraphics[width=0.9\textwidth]{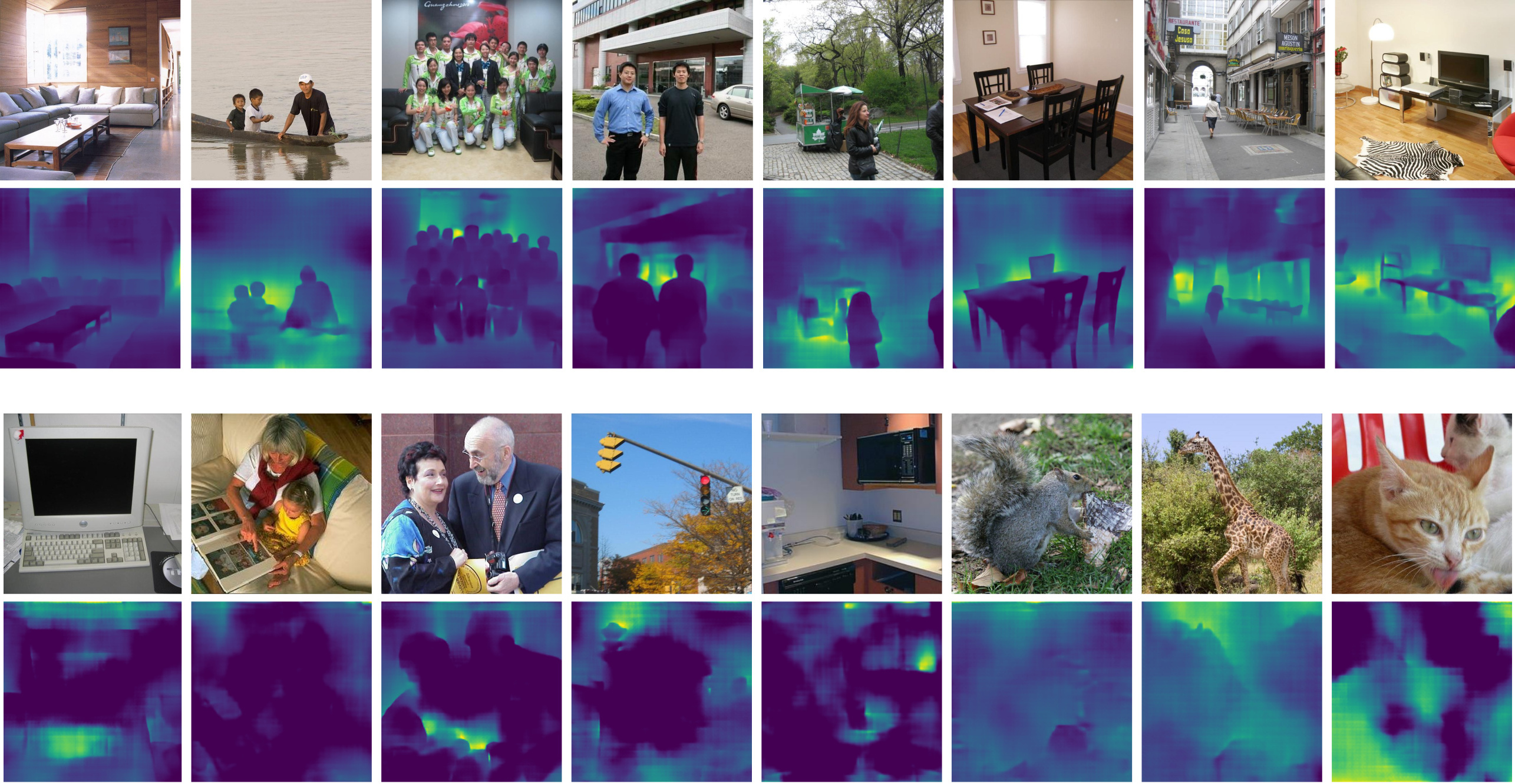}
      \end{center}
  \caption{The first two rows are the examples of visual genome images and their synthetically generated high quality depth maps. The second two rows are the examples of visual genome images and their synthetically generated noisy depth maps. }\label{goodones}
\end{figure*}

\section{Conclusion}
We employed an RGB-to-Depth network, trained on a large corpus of data, to generate depth maps for Visual Genome dataset, releasing a new extension called \textit{VG-Depth}. We provided a metric, \textit{Macro R@K} for better evaluation of relation detection in Visual Genome and other highly imbalanced datasets. In extensive empirical evaluations, we demonstrated the effect of different object features in visual relation detection and showed that by using depth information, we achieve significantly better performance compared to other state-of-the-art methods.

\section{Acknowledgements}
We thank Evgeniy Faerman, Vaheh Hatami, Alireza Ghazaei and the anonymous reviewers for their fruitful comments. This work was supported by the BMBF as part of the project MLWin (01IS18050).

\bibliographystyle{IEEEtran}
\bibliography{icpr2020}

\end{document}